\documentclass[conference]{IEEEtran}
\IEEEoverridecommandlockouts
\usepackage{cite}
\usepackage{amsmath,amssymb,amsfonts}
\usepackage{algorithm,algpseudocode}
\usepackage{graphicx}
\usepackage{textcomp}
\usepackage{xcolor}
\usepackage{hyperref}
\usepackage[flushleft]{threeparttable}

\def\BibTeX{{\rm B\kern-.05em{\sc i\kern-.025em b}\kern-.08em
    T\kern-.1667em\lower.7ex\hbox{E}\kern-.125emX}}
\begin{document}

\makeatletter
\newcommand{\linebreakand}{%
\end{@IEEEauthorhalign}
\hfill\mbox{}\par
\mbox{}\hfill\begin{@IEEEauthorhalign}
}
\makeatother

\title{Event-Driven Tactile Learning with \\Location Spiking Neurons}


\author{\IEEEauthorblockN{Peng Kang\IEEEauthorrefmark{1},
		Srutarshi Banerjee\IEEEauthorrefmark{2}, Henry Chopp\IEEEauthorrefmark{2}, 
		Aggelos Katsaggelos\IEEEauthorrefmark{1}\IEEEauthorrefmark{2}, Oliver Cossairt\IEEEauthorrefmark{1}\IEEEauthorrefmark{2}}
	\IEEEauthorblockA{\IEEEauthorrefmark{1}Department of Computer Science,
		Northwestern University, Evanston, IL 60208, USA \\
		\IEEEauthorrefmark{2}Department of Electrical and Computer Engineering,
		Northwestern University, Evanston, IL 60208, USA\\
		Email: \{pengkang2022, SrutarshiBanerjee2022, henrychopp2017\}@u.northwestern.edu,\\
		\{a-katsaggelos, oliver.cossairt\}@northwestern.edu}}
	
%

\maketitle

\begin{abstract}
The sense of touch is essential for a variety of daily tasks. New advances in event-based tactile sensors and Spiking Neural Networks (SNNs) spur the research in event-driven tactile learning. However, SNN-enabled event-driven tactile learning is still in its infancy due to the limited representative abilities of existing spiking neurons and high spatio-temporal complexity in the data. In this paper, to improve the representative capabilities of existing spiking neurons, we propose a novel neuron model called ``location spiking neuron'', which enables us to extract features of event-based data in a novel way. Moreover, based on the classical Time Spike Response Model (TSRM), we develop a specific location spiking neuron model -- Location Spike Response Model (LSRM) that serves as a new building block of SNNs\footnote{The TSRM is the classical SRM in the literature. We add the character ``T'' to highlight its difference with the LSRM.}. Furthermore, we propose a hybrid model which combines an SNN with TSRM neurons and an SNN with LSRM neurons to capture the complex spatio-temporal dependencies in the data. Extensive experiments demonstrate the significant improvements of our models over other works on event-driven tactile learning and show the superior energy efficiency of our models and location spiking neurons, which may unlock their potential on neuromorphic hardware.\footnote{Copyright © 2022 IEEE. Personal use of this material is permitted. Permission from IEEE must be obtained for all other uses, in any current or future media, including reprinting/republishing this material for advertising or promotional purposes, creating new collective works, for resale or redistribution to servers or lists, or reuse of any copyrighted component of this work in other works by sending a request to pubs-permissions@ieee.org.}
\end{abstract}


%
%
%

\section{Introduction}
The tactile perception is a vital sensing modality that enables humans to gain perceptual judgment on the surrounding environment and conduct stable movement~\cite{taunyazov2020fast}. 
With the recent advances in material science and Artificial Neural Networks (ANNs), research on tactile perception begins to soar, including tactile object recognition~\cite{soh2014incrementally, kappassov2015tactile, sanchez2018online}, slip detection~\cite{calandra2018more}, and texture recognition~\cite{baishya2016robust, taunyazov2019towards}. Unfortunately, although ANNs demonstrate promising performance on the tactile learning tasks, they are usually power-hungry compared to human brains that require far less energy to perform the tactile perception robustly~\cite{li2016evaluating, strubell2019energy}. 

Inspired by biological systems, research on event-driven perception starts to gain momentum, and several asynchronous event-based sensors have been proposed, including event cameras~\cite{gallego2020event} and event-based tactile sensors~\cite{taunyazov2020event}. In contrast to standard synchronous sensors, such event-based sensors can achieve higher energy efficiency, better scalability, and lower latency. However, due to the high sparsity and complexity of event-driven data, learning with these sensors remains in its infancy~\cite{pfeiffer2018deep}. Recently, several works~\cite{taunyazov2020event, gu2020tactilesgnet, taunyazov2020fast} utilized Spiking Neural Networks (SNNs)~\cite{Shrestha2018, pfeiffer2018deep, ijcai2021-441} to tackle event-driven tactile learning. Unlike ANNs generally requiring expensive transformations from asynchronous discrete events to synchronous real-valued frames, SNNs can process event-based sensor data directly. Moreover, unlike ANNs employing artificial neurons~\cite{maas2013rectifier, xu2015empirical, clevert2015fast} and conducting real-valued computation, SNNs adopt spiking neurons~\cite{gerstner1995time, abbott1999lapicque, gerstner2002spiking} and utilize binary 0-1 spikes to process information. This difference reduces the mathematical dot-product operations in ANNs to less computationally summation operations in SNNs. Due to the advantages of SNNs, these works are always energy-efficient and suitable for power-constrained devices. However, due to the limited representative abilities of current spiking neuron models and high spatio-temporal complexity in the event-based tactile data, these works still cannot sufficiently capture spatio-temporal dependencies and thus hinder the performance of event-driven tactile learning.

In this paper, to address the problems mentioned above, we make several contributions that advance event-driven tactile learning. 

First, to enable richer representative abilities of existing spiking neurons, we propose a novel neuron model called ``location spiking neuron''. Unlike existing spiking neuron models that update their membrane potentials based on time steps~\cite{roy2019towards}, location spiking neurons update their membrane potentials based on locations. Moreover, based on the Time Spike Response Model (TSRM)~\cite{gerstner1995time}, we develop the ``Location Spike Response Model'', henceforth referred to as ``LSRM''. The TSRM is the classical SRM in the literature. We add the character ``T (Time)'' to highlight its difference with the LSRM. These location spiking neurons enable us to extract feature representations of event-based data in a novel way. Previously, SNNs adopted temporal recurrent neuronal dynamics to extract features from the event-based data. With location spiking neurons, we can build SNNs that employ spatial recurrent neuronal dynamics to extract features from the event-based data. We believe location spiking neuron models can have a broad impact on the SNN community and spur the research on learning from event sensors like NeuTouch~\cite{taunyazov2020event} or Dynamic Vision Sensors~\cite{gallego2020event}.

Next, we investigate the effectiveness of location spiking neurons and develop a hybrid model to capture the complex spatio-temporal dependencies in the event-driven data. The hybrid model combines an SNN with TSRM neurons and an SNN with LSRM neurons. Moreover, we introduce a location spike-count loss and a weighted spike-count loss to train the SNN with LSRM neurons and the hybrid model, respectively. 

Last but not least, we apply our proposed models to event-driven tactile learning, including event-driven tactile object recognition and event-driven slip detection, and test them on three challenging datasets. Specifically, two sub-tasks are included in the task of event-driven tactile object recognition. The first sub-task requires models to determine the type of objects being handled. The second sub-task requires models to determine the type of containers being handled and the amount of liquid held within, which is more challenging than the first sub-task. In the task of event-driven slip detection, models need to accurately detect the rotational slip (``stable'' or ``rotate'') within 0.15s. Extensive experimental results demonstrate the significant improvements of our models over other state-of-the-art methods on event-driven tactile learning and show the superior energy efficiency of our models, which may bring new opportunities and unlock their potential on neuromorphic hardware. 

To the best of our knowledge, this is the first work to propose location spiking neurons, introduce the LSRM, and build SNNs with location spiking neurons for event-driven tactile learning. 
The rest of the paper is organized as follows. In Section~\ref{sec:related}, we give an overview of related work on SNNs and event-driven tactile sensing and learning. In Section~\ref{sec:methods}, we start by introducing notations for TSRM neurons and extending them to the specific location spiking neurons -- LSRM neurons. We then propose models with LSRM neurons for event-driven tactile learning. Last, we provide implementation details and algorithms related to the proposed models. In Section~\ref{sec:exp}, we demonstrate the effectiveness and energy efficiency of our proposed models on different benchmark datasets. Finally, we discuss the broad impact of this work and conclude in Section~\ref{sec:conclude}

\section{Related Work}\label{sec:related}
In the following, we give a brief overview of related work on SNNs and event-driven tactile sensing and learning.
\subsection{Spiking Neural Networks (SNNs)}
With the prevalence of Artificial Neural Networks (ANNs), computers today have demonstrated extraordinary abilities in many cognition tasks. However, ANNs only imitate brain structures in several ways, including vast connectivity and structural and functional organizational hierarchy \cite{roy2019towards}. The brain has more information processing mechanisms like the neuronal and synaptic functionality \cite{bullmore2012economy, felleman1991distributed}. Moreover, ANNs are much more energy-consuming than human brains. To integrate more brain-like characteristics and make artificial intelligence models more energy-efficient, researchers propose Spiking Neural Networks (SNNs), which can be executed on power-efficient neuromorphic processors like TrueNorth~\cite{merolla2014million} and Loihi~\cite{davies2021advancing}. Similar to ANNs, SNNs can adopt general network topologies like convolutional layers and fully-connected layers, but use different neuron models~\cite{gerstner2002spiking}. Commonly-used neuron models for SNNs are the Leaky Integrate-and-Fire (LIF) model~\cite{abbott1999lapicque} and the Time Spike Response Model (TSRM)~\cite{gerstner1995time}. Due to the non-differentiability of these spiking neuron models, it still remains challenging to train SNNs. Nevertheless, several solutions have been proposed, such as converting trained ANNs to SNNs~\cite{cao2015spiking,sengupta2019going} and approximating the derivative of the spike function~\cite{wu2018spatio, ijcai2020-211}. In this work, we propose location spiking neurons to enhance the representative abilities of existing spiking neurons. These location spiking neurons maintain the spiking characteristic but employ the spatial recurrent neuronal dynamics, which enable us to build energy-efficient SNNs and extract features of event-based data in a novel way. Moreover, based on the optimization methods for SNNs with existing spiking neurons, we can derive the approximate backpropagation methods for SNNs with location spiking neurons.
\subsection{Event-Driven Tactile Sensing and Learning}
With the prevalence of material science and robotics, several tactile sensors have been developed, including non-event-based tactile sensors like the iCub RoboSkin~\cite{schmitz2010tactile} and the SynTouch BioTac\cite{fishel2012sensing} and event-driven tactile sensors like the NeuTouch~\cite{taunyazov2020event} and the NUSkin~\cite{taunyazov2021extended}. In this paper, we focus on event-driven tactile learning with SNNs. Since the development of event-driven tactile sensors is still in its infancy~\cite{gu2020tactilesgnet}, little prior work exists on learning event-based tactile data with SNNs. The work~\cite{taunyazov2020fast} employed a neural coding scheme to convert raw tactile data from non-event-based tactile sensors into event-based spike trains. It then utilized an SNN to process the spike trains and classify textures. A recent work~\cite{taunyazov2020event} released the first publicly-available event-driven visual-tactile dataset collected by NeuTouch and proposed an SNN based on SLAYER~\cite{Shrestha2018} to solve the event-driven tactile learning. Moreover, to naturally capture the spatial topological relations and structural knowledge in the event-based tactile data, a very recent work~\cite{gu2020tactilesgnet} adopted the spiking graph neural network~\cite{ijcai2021-441} to process the event-based tactile data and conduct the tactile object recognition. In this paper, different from previous works building SNNs with spiking neurons that employ the temporal recurrent neuronal dynamics, we construct SNNs with location spiking neurons to capture the complex spatio-temporal dependencies in the event-based tactile data and boost the event-driven tactile learning.

\section{Methods}\label{sec:methods}
In this section, we first demonstrate the spatial recurrent neuronal dynamics of location spiking neurons by introducing notations for the existing spiking neuron model -- TSRM and extending it to the location spiking neuron model -- LSRM. We then introduce models with location spiking neurons for event-driven tactile learning. Last, we provide implementation details and algorithms related to the proposed models.
\subsection{Location Spiking Neurons}
Spiking neuron models are mathematical descriptions of specific cells in the nervous system. They are the basic building blocks of SNNs. Two commonly-used spiking neuron models are the LIF model and the TSRM. Since the TSRM is more general than the LIF model~\cite{maass2001pulsed}, we introduce the TSRM and transform it to a location spiking neuron model -- the LSRM. A similar transformation process can be applied to the LIF model to derive its corresponding location spiking neuron model. 
\begin{figure}
	\includegraphics[width=\linewidth]{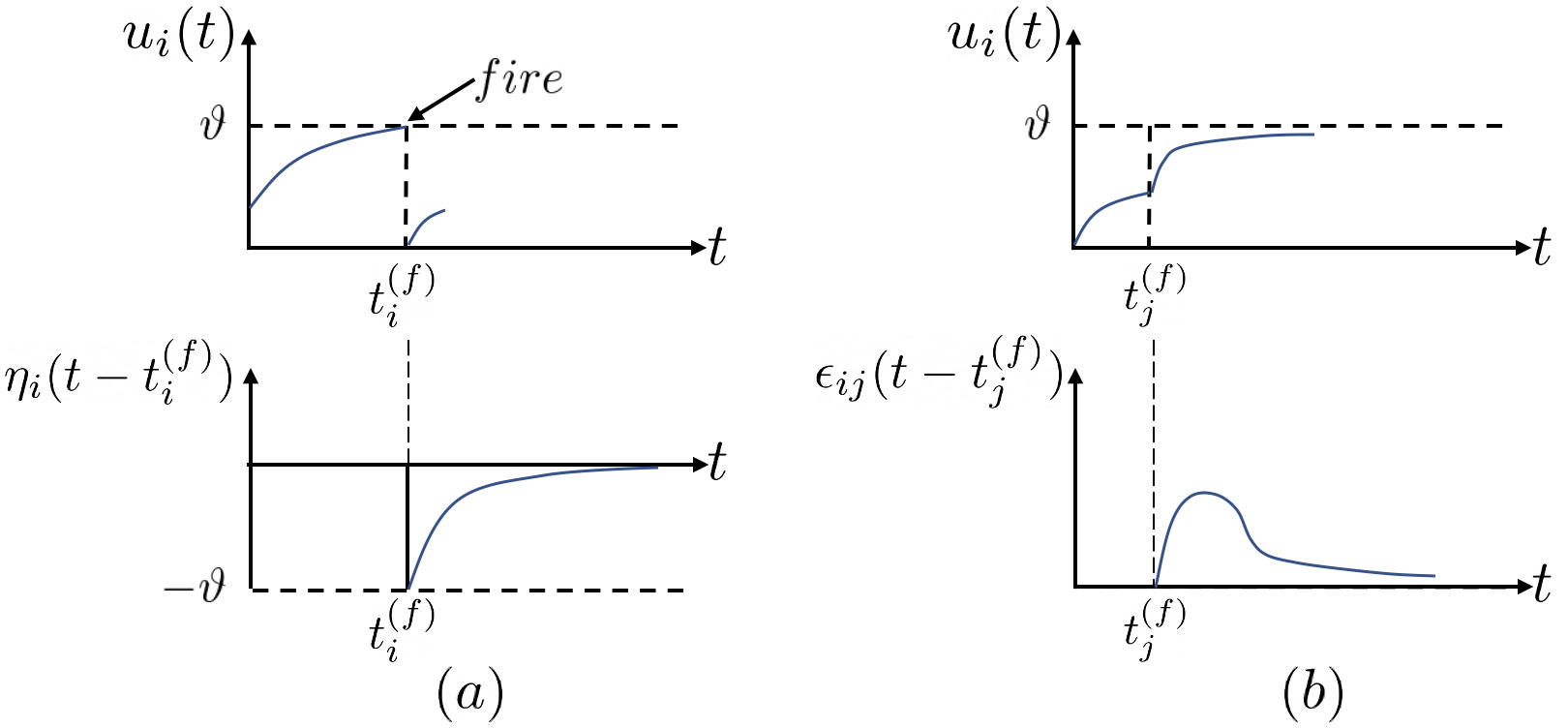}
	\caption{The \textbf{temporal} recurrent neuronal dynamics of TSRM neuron $i$. (a) the refractory dynamics of TSRM neuron $i$. Immediately after firing an output spike at time $t_i^{(f)}$, the value of $u_i(t)$ is lowered or reset by adding a negative contribution $\eta_i(\cdot)$. The kernel $\eta_i(\cdot)$ vanishes for $t<t_i^{(f)}$ and decays to zero for $t\to\infty$. (b) the incoming spike dynamics of TSRM neuron $i$. A presynaptic spike at time $t_j^{(f)}$ increases the value of $u_i(t)$ for $t\geq t_j^{(f)}$ by an amount of $w_{ij}x_j(t_j^{(f)})\epsilon_{ij}(t-t_j^{(f)})$. The kernel $\epsilon_{ij}(\cdot)$ vanishes for $t<t_j^{(f)}$. A transmission delay may be included in the definition of $\epsilon_{ij}(\cdot)$.}
	\label{temporalDynamics}
\end{figure}

In the TSRM, the temporal recurrent neuronal dynamics of neuron $i$ are described by its membrane potential $u_i(t)$. When $u_i(t)$ exceeds a predefined threshold $\vartheta$ at the firing time $t_i^{(f)}$, the neuron $i$ will generate a spike. The set of all firing times of neuron $i$ is denoted by 
\begin{equation}
	\label{e1}
	\mathcal{F}_i=\{t_i^{(f)}; 1\leq f\leq n\} = \{t|u_i(t)=\vartheta\}, 
\end{equation}  
where $t_i^{(n)}$ is the most recent spike time $t_i^{(f)}<t$. The value of $u_i(t)$ is governed by two different spike response processes:
\begin{equation}
	\label{e2}
	u_i(t) = \sum_{t_i^{(f)}\in\mathcal{F}_i}\eta_i(s_i) + \sum_{j\in\Gamma_i}\sum_{t_j^{(f)}\in\mathcal{F}_j}w_{ij}x_j(t_j^{(f)})\epsilon_{ij}(s_j), 
\end{equation}  
where $s_i=t-t_i^{(f)}$, $s_j=t-t_j^{(f)}$, $\Gamma_i$ is the set of presynaptic neurons of neuron $i$, and $x_j(t_j^{(f)})=1$ is the presynaptic spike. $\eta_i(\cdot)$ is the refractory kernel, which describes the response of neuron $i$ to its own spikes at time $t$. $\epsilon_{ij}(\cdot)$ is the incoming spike response kernel, which models the neuron $i$'s response to the presynaptic spikes from neuron $j$ at time $t$. $w_{ij}$ accounts for the connection strength between neuron $i$ and neuron $j$ and scale the incoming spike response. Figure~\ref{temporalDynamics}(a) visualizes the refractory dynamics and Figure~\ref{temporalDynamics}(b) visualizes the incoming spike dynamics. Without loss of generality, such temporal recurrent neuronal dynamics also apply to other spiking neuron models, such as LIF neurons. From the above descriptions, we find that existing spiking neuron models \textbf{explicitly} convolve the temporal information in the data but fail to \textbf{explicitly} convolve the spatial information in the data, which, to some extent, limits their representative abilities. 
\begin{figure}
	\includegraphics[width=\linewidth]{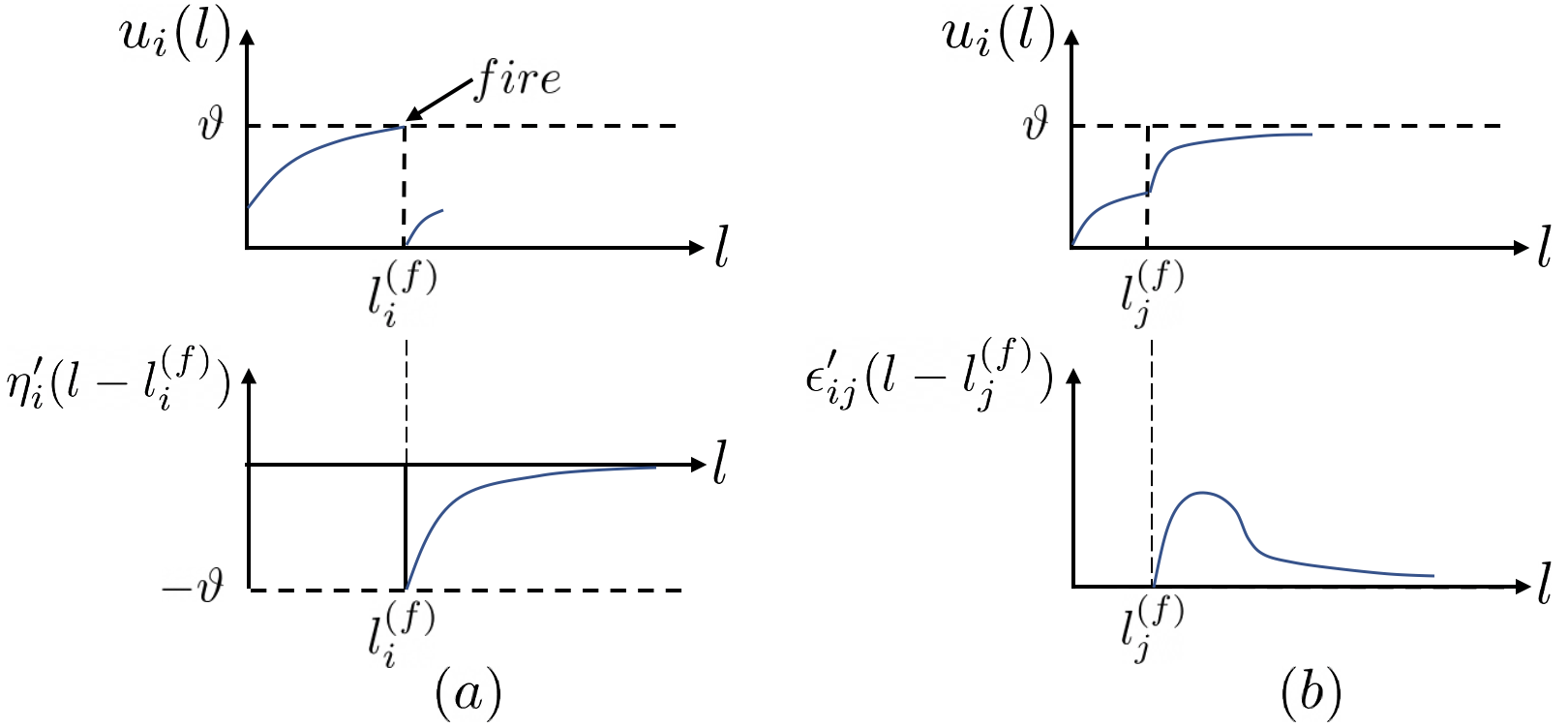}
	\caption{The \textbf{spatial} recurrent neuronal dynamics of LSRM neuron $i$. (a) the refractory dynamics of LSRM neuron $i$. Immediately after firing an output spike at location $l_i^{(f)}$, the value of $u_i(l)$ is lowered or reset by adding a negative contribution $\eta'_i(\cdot)$. The kernel $\eta'_i(\cdot)$ vanishes for $l<l_i^{(f)}$ and decays to zero for $l\to\infty$. (b) the incoming spike dynamics of LSRM neuron $i$. A presynaptic spike at location $l_j^{(f)}$ increases the value of $u_i(l)$ for $l\geq l_j^{(f)}$ by an amount of $w'_{ij}x'_j(l_j^{(f)})\epsilon'_{ij}(l-l_j^{(f)})$. The kernel $\epsilon'_{ij}(\cdot)$ vanishes for $l<l_j^{(f)}$. A location delay may be included in the definition of $\epsilon'_{ij}(\cdot)$. ``$<$'' and ``$\geq$'' indicate the location order.}
	\label{spatialDynamics}
\end{figure}

\begin{figure*}
	\includegraphics[width=\linewidth]{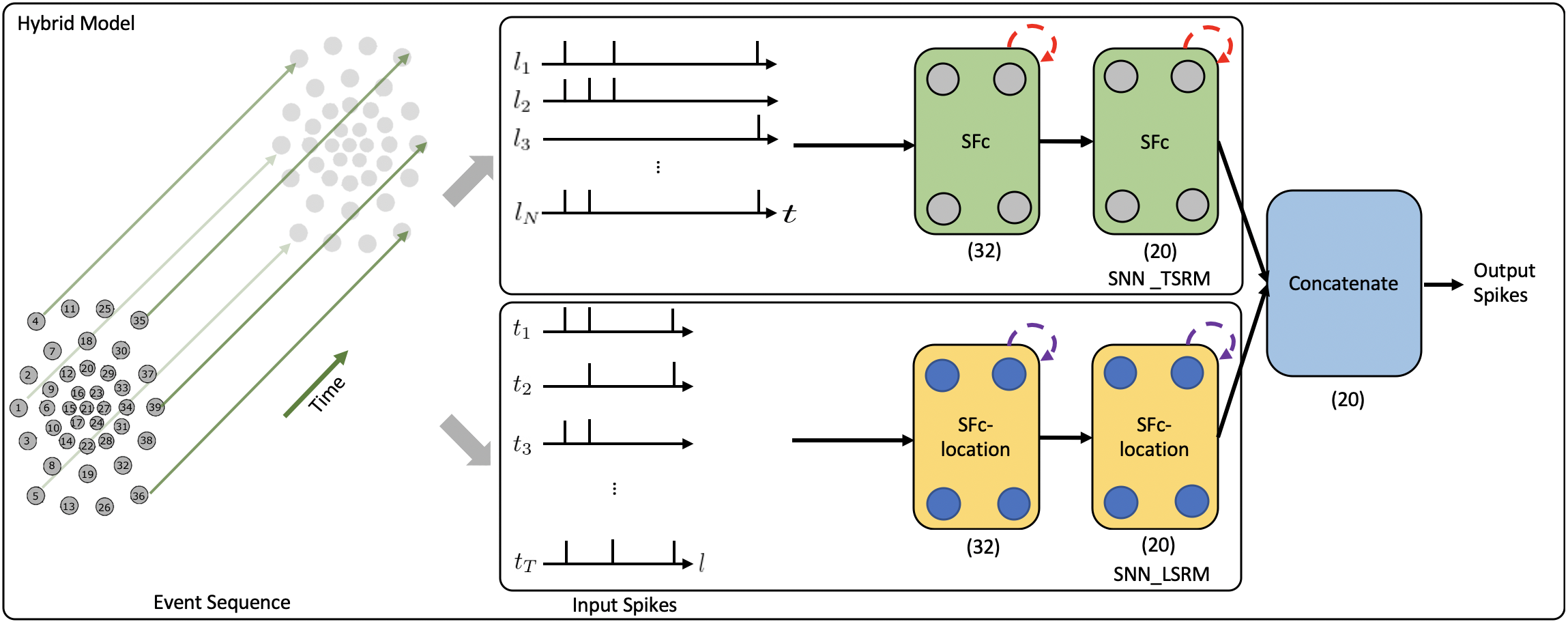}
	\caption{The network structure of the hybrid model. The SNN with TSRM neurons (SNN\_TSRM) processes the input spikes $X_{in}$ and adopts the temporal recurrent neuronal dynamics (shown with red dashed arrows) of TSRM neurons to extract features from the data. The SNN with LSRM neurons (SNN\_LSRM) processes the transposed input spikes $X'_{in}$ and employs the spatial recurrent neuronal dynamics (shown with purple dashed arrows) of LSRM neurons to extract features from the data. Finally, the spiking representations from two networks are concatenated to yield the final predicted label. (32) and (20) represent the sizes of fully-connected layers, where we assume the number of classes ($K$) to be equal to 20.}
	\label{hybrid}
\end{figure*}
To enrich the representative abilities of existing spiking neuron models, we propose location spiking neurons, which adopt the spatial recurrent neuronal dynamics and update their membrane potentials based on locations\footnote{locations could refer to pixel locations for images or taxel locations for tactile sensors.}. These neurons are able to \textbf{explicitly} convolve the spatial information in the data and enable us to extract features of event-based data in a novel way. Specifically, we adopt the TSRM and transform it to the LSRM. In the LSRM, the spatial recurrent neuronal dynamics of neuron $i$ are described by its location membrane potential $u_i(l)$. When $u_i(l)$ exceeds a predefined threshold $\vartheta$ at the firing location $l_i^{(f)}$, the neuron $i$ will generate a spike. The set of all firing locations of neuron $i$ is denoted by 
\begin{equation}
	\label{e3}
	\mathcal{G}_i=\{l_i^{(f)}; 1\leq f\leq n\} = \{l|u_i(l)=\vartheta\}, 
\end{equation}  
where $l_i^{(n)}$ is the nearest firing location $l_i^{(f)} < l$. ``$<$'' indicates the location order, which is manually set and will be discussed in Section~\ref{locationOrder}. The value of $u_i(l)$ is governed by two different spike response processes:
\begin{equation}
	\label{e4}
	u_i(l) = \sum_{l_i^{(f)}\in\mathcal{G}_i}\eta'_i(s'_i) + \sum_{j\in\Gamma'_i}\sum_{l_j^{(f)}\in\mathcal{G}_j}w'_{ij}x'_j(l_j^{(f)})\epsilon'_{ij}(s'_j), 
\end{equation}  
where $s'_i=l-l_i^{(f)}$, $s'_j=l-l_j^{(f)}$, $\Gamma'_i$ is the set of presynaptic neurons of neuron $i$, and $x'_j(l_j^{(f)})=1$ is the presynaptic spike. $\eta'_i(\cdot)$ is the refractory kernel, which describes the response of neuron $i$ to its own spikes at location $l$. $\epsilon'_{ij}(\cdot)$ is the incoming spike response kernel, which models the neuron $i$'s response to the presynaptic spikes from neuron $j$ at location $l$. $w_{ij}$ accounts for the connection strength between neuron $i$ and neuron $j$ and scale the incoming spike response. Figure~\ref{spatialDynamics}(a) visualizes the refractory dynamics of LSRM neurons and Figure~\ref{spatialDynamics}(b) visualizes the incoming spike dynamics of LSRM neurons. The threshold $\vartheta$ of LSRM neurons can be different from that of TSRM neurons, while we set the same for simplicity.

\subsection{Event-Driven Tactile Learning with Location Spiking Neurons}
Such location spiking neurons enable us to extract feature representations of event-based data in a novel way. To take advantage of location spiking neurons and boost the event-based tactile learning performance, we propose models with location spiking neurons, which capture complex spatio-temporal dependencies in the event-based tactile data. In this paper, we focus on processing the data collected by NeuTouch~\cite{taunyazov2020event}, a biologically-inspired event-driven fingertip tactile sensor with 39 taxels arranged spatially in a radial fashion (Fig.~\ref{hybrid}).
\subsubsection{Hybrid Model}\label{sec:snn}
Figure~\ref{hybrid} presents the network structure of the hybrid model. From the figure, we can see that the hybrid model has two components, including the SNN with TSRM neurons (SNN\_TSRM) and the SNN with LSRM neurons (SNN\_LSRM). Specifically, SNN\_TSRM employs the temporal recurrent neuronal dynamics to extract spiking feature representations from the event-based tactile data $X_{in} \in \mathbb{R}^{N\times T}$, where $N$ is the total number of taxels and $T$ is the total time length of event sequences. SNN\_LSRM utilizes the spatial recurrent neuronal dynamics to extract spiking feature representations from the event-based tactile data $X'_{in} \in \mathbb{R}^{T\times N}$, where $X'_{in}$ is transposed from $X_{in}$. The spiking representations from two networks are then concatenated to yield the final task-specific output.
\begin{figure}
	\includegraphics[width=\linewidth]{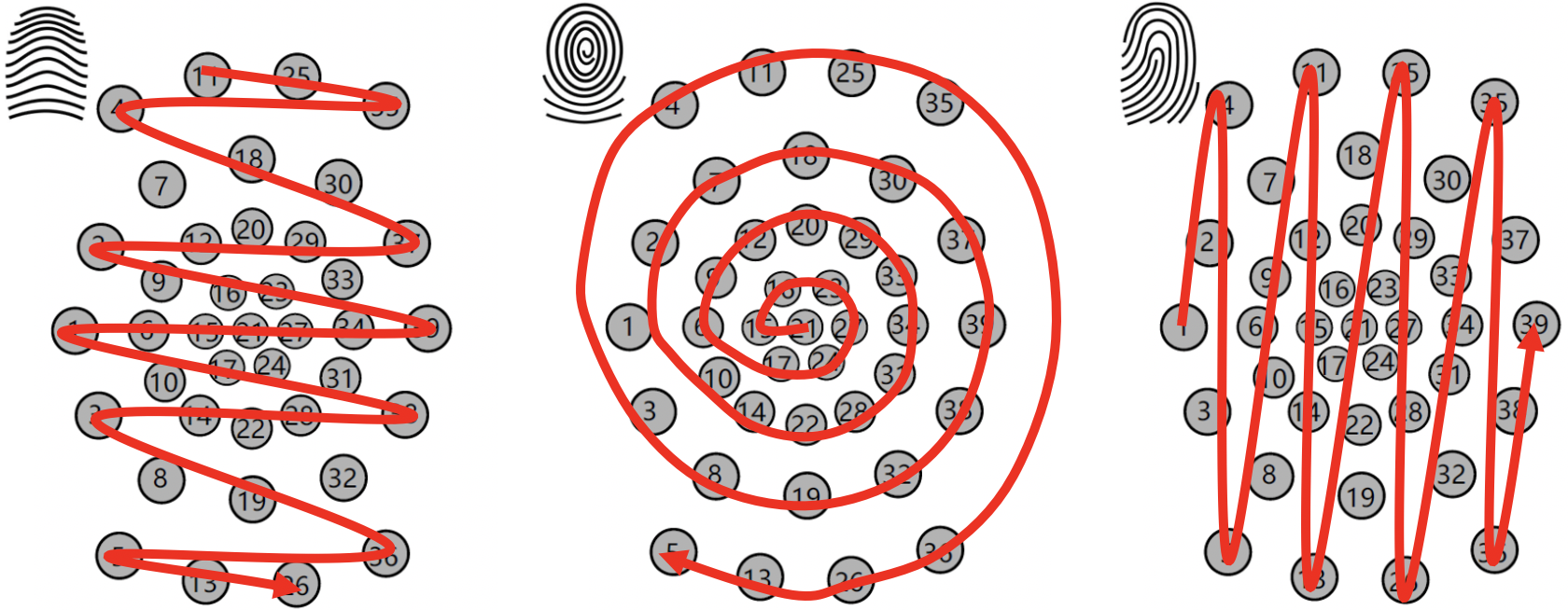}
	\caption{Three bio-inspired location orders. Left to right: arch-like location order, whorl-like location order, loop-like location order.}
	\label{orders}
\end{figure}
\subsubsection{SNN\_TSRM vs.\ SNN\_LSRM}
The network structure of SNN\_TSRM is shown in the top part of Fig.~\ref{hybrid}. It employs two spiking fully-connected layers with TSRM neurons (SFc) to process $X_{in}$ and generate the spiking representations $O_{1}\in\mathbb{R}^{K\times T}$, where $K$ is the output dimension determined by the task. The membrane potential $u_i(t)$, the output spiking state $o_i(t)$, and the set of all firing times $\mathcal{F}_i $ of TSRM neuron $i$ in SFc are decided by:
\begin{equation}	\label{e5}
	\begin{split}
		u_i(t) &= \sum_{t_i^{(f)}\in\mathcal{F}_i}\eta(s_i) + \underbrace{\sum_{j\in\Gamma_i}\sum_{t_j^{(f)}\in\mathcal{F}_j}w_{ij}o_j(t_j^{(f)})\epsilon(s_j)}_\text{capture spatial dependencies},   \\
		o_i(t) &= \begin{cases}
			1 & \text{if $u_i(t) > \vartheta$};\\
			0 & \text{otherwise},
		\end{cases}  \\
	    \mathcal{F}_i &= \begin{cases}
		     \mathcal{F}_i \cup t & \text{if $o_i(t) = 1$};\\
		     \mathcal{F}_i & \text{otherwise},
		\end{cases}           
	\end{split}
\end{equation}
where $w_{ij}$ are the trainable parameters, $\eta$($\cdot$) and $\epsilon$($\cdot$) are predefined by hyperparameters, $\Gamma_i$ is the set of presynaptic neurons spanning over the spatial domain, which is utilized to capture the spatial dependencies in the event-based data.

The network structure of SNN\_LSRM is shown in the bottom part of Fig.~\ref{hybrid}. It employs two spiking fully-connected layers with LSRM neurons (SFc-location) to process $X'_{in}$ and generate the spiking representations $O_{2}\in\mathbb{R}^{K\times N}$, where $K$ is the output dimension decided by the task. The membrane potential $u_i(l)$, the output spiking state $o_i(l)$, and the set of all firing locations $\mathcal{G}_i $ of LSRM neuron $i$ in SFc-location are decided by:
\begin{equation}	\label{e6}
	\begin{split}
		u_i(l) &= \sum_{l_i^{(f)}\in\mathcal{G}_i}\eta'(s'_i) + \underbrace{\sum_{j\in\Gamma'_i}\sum_{l_j^{(f)}\in\mathcal{G}_j}w'_{ij}o_j(l_j^{(f)})\epsilon'(s'_j)}_\text{model temporal dependencies},   \\
		o_i(l) &= \begin{cases}
			1 & \text{if $u_i(l) > \vartheta$};\\
			0 & \text{otherwise},
		\end{cases}   \\
		 \mathcal{G}_i &= \begin{cases}
			\mathcal{G}_i \cup l & \text{if $o_i(l) = 1$};\\
			\mathcal{G}_i & \text{otherwise},
		\end{cases}           
	\end{split}    
\end{equation}
where $w'_{ij}$ are the trainable parameters, $\eta'$($\cdot$) and $\epsilon'$($\cdot$) are predefined by hyperparameters, $\Gamma'_i$ is the set of presynaptic neurons spanning over the temporal domain, which is utilized to model the temporal dependencies in the event-based data. Such location spiking neurons tap the representative potential and enable us to capture features in this way.

\subsubsection{Concatenate}
We concatenate the spiking representations of $O_{1}$ and $O_{2}$ along the last dimension and obtain the final output spike train $O\in\mathbb{R}^{K\times (T+N)}$. The predicted label is associated with the neuron $k\in K$ with the largest number of spikes in the duration of $T+N$. 
\subsubsection{Location Orders}\label{locationOrder}
To enable location spiking neurons' spatial recurrent neuronal dynamics, we propose three location orders for event-based tactile learning (Fig.~\ref{orders}) based on three major fingerprint patterns of humans -- arch, whorl, and loop. 
Three examples are shown here. Each number in the brackets represents the taxel index shown in Fig.~\ref{hybrid}.
\begin{itemize}
\item An example for the arch-like location order: [11, 25, 35, 4, 18, 30, 7, 2, 20, 37, 29, 12, 9, 33, 23, 16, 1, 6, 15, 21, 27, 34, 39, 24, 17, 10, 31, 38, 28, 14, 3, 22, 32, 8, 19, 36, 5, 13, 26] 
\item An example for the whorl-like location order: [21, 15, 16, 23, 27, 24, 17, 6, 9, 12, 20, 29, 33, 34, 31, 28, 22, 14, 10, 1, 2, 7, 18, 30, 37, 39, 38, 32, 19, 8, 3, 4, 11, 25, 35, 36, 26, 13, 5]
\item An example for the loop-like location order: [1, 2, 3, 4, 5, 6, 7, 8, 9, 10, 11, 12, 13, 14, 15, 16, 17, 18, 19, 20, 21, 22, 23, 24, 25, 26, 27, 28, 29, 30, 31, 32, 33, 34, 35, 36, 37, 38, 39] 
\end{itemize}
\subsection{Implementation Details and Algorithms}\label{sec:alg}
Similar to the spike-count loss of prior works~\cite{Shrestha2018, taunyazov2020event}, we propose a location spike-count loss to optimize the SNN with LSRM neurons:
\begin{equation}
	\label{e7}
	\mathcal{L}_{location}  =\frac{1}{2}\sum_{k=0}^{K}\left(\sum_{l=0}^{N}o_{k}(l) - \sum_{l=0}^{N}\hat{o}_{k}(l)\right)^2, 
\end{equation}  
which captures the difference between the observed output spike count $\sum_{l=0}^{N}o_{k}(l)$ and the desired spike count $\sum_{l=0}^{N}\hat{o}_{k}(l)$ across the $K$ neurons. Moreover, to optimize the hybrid model, we develop a weighted spike-count loss:
\begin{equation}
	\label{e8}
	\mathcal{L} = \frac{1}{2}\sum_{k=0}^{K}\left(\sum_{t=0}^{T}o_{k}(t) + \lambda\sum_{l=0}^{N}o_{k}(l) - \sum_{c=0}^{T+ N}\hat{o}_{k}(c)\right)^2, 
\end{equation}  
which first balances the contributions from two SNNs and then captures the difference between the observed balanced output spike count $\sum_{t=0}^{T}o_{k}(t) + \lambda\sum_{l=0}^{N}o_{k}(l)$ and the desired spike count $\sum_{c=0}^{T+ N}\hat{o}_{k}(c)$ across the $K$ output neurons. For both $\mathcal{L}_{location}$ and $\mathcal{L}$, the desired spike counts have to be specified for the correct and incorrect classes and are task-dependent hyperparameters. We set these hyperparameters like~\cite{taunyazov2020event} for simplicity. To overcome the non-differentiability of spikes and apply the backpropagation algorithm, we use the approximate gradient proposed in SLAYER~\cite{Shrestha2018}. The timestep-wise inference algorithm of the hybrid model is shown in Alg.~\ref{alg:timestep}. And the corresponding timestep-wise training algorithm can be derived by incorporating the weighted spike-count loss.
\begin{algorithm}
	\caption{Timestep-wise inference algorithm}
	\label{alg:timestep}
	\begin{algorithmic}[1]
		\Require{event-based tactile inputs $X_{in} \in \mathbb{R}^{N\times T}$, $N$ taxels, and the total time length $T$.} 
		\Ensure{timestep-wise predictions of $O_{1}$, $O_{2}$, and $O$.}
		\For{$t \gets 1$ to $T$}     
		\State {obtain $X\in \mathbb{R}^{N\times t}$}  
		\State {obtain $\bar{X'}=concatenate(X', \mathbf{0}) \in\mathbb{R}^{T\times N}$, where $X'\in\mathbb{R}^{t\times N}$, and $\mathbf{0}\in\mathbb{R}^{(T-t)\times N}$}  
		\State {$O_{1}(t)=\mathbf{0}\in\mathbb{R}^{K\times t}$, $O_{2}(t)=\mathbf{0}\in\mathbb{R}^{K\times N}$}  
		\State {$O(t)=\mathbf{0}\in\mathbb{R}^{K\times (t+N)}$}  
		\State {$O_{1}(t)$ = SNN\_TSRM($X$)}
		\State {$O_{2}(t)$ = SNN\_LSRM($\bar{X'}$)}
		\State {$O(t)$ = $concatenate$($O_{1}(t), O_{2}(t)$) }
		\EndFor              
	\end{algorithmic}
\end{algorithm}

%

\begin{figure*}
	\includegraphics[width=\linewidth]{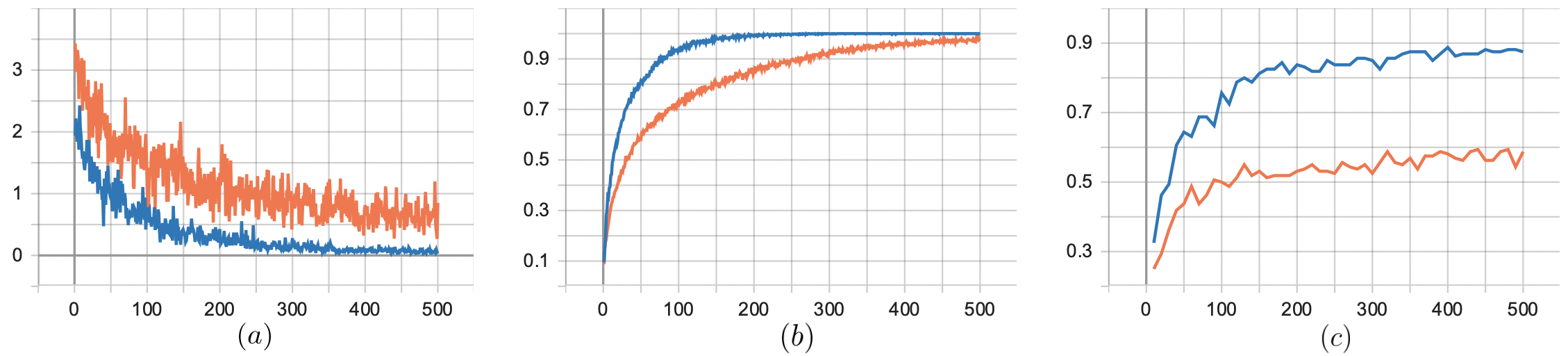}
	\caption{Training and testing profiles for Tactile-SNN (orange) and Ours-Hybrid (blue): (a) the training loss, (b) the training accuracy, (c) the testing accuracy.}
	\label{profile}
\end{figure*}
\section{Experiments}\label{sec:exp}
In this section, we first introduce the datasets and models for event-driven tactile learning. Next, to show the effectiveness of our proposed models, we extensively evaluate their performance on three event-driven tactile datasets and compare them with state-of-the-art models. Finally, we demonstrate the superior energy efficiency of our proposed models over ANNs and show the high-efficiency benefit of location spiking neurons. We utilize the slayerPytorch framework~\footnote{https://github.com/bamsumit/slayerPytorch} to implement the proposed models and employ RMSProp with the $l_2$ regularization to optimize them. The source code is available at \url{https://github.com/pkang2017/TactileLocNeurons}.
\begin{table}[ht]
	\caption{Dataset statistics}
	\setlength{\tabcolsep}{2.4mm}{
		\begin{tabular}{lcccccc}
			\hline
			Datasets       & TD (s) & SSR (s) & $T$   & $N$  & $K$ & \#Samples \\ \hline
			Objects        & 6.5               & 0.02                    & 325 & 78 &  36        & 900       \\ \hline
			Containers     & 6.5               & 0.02                   & 325 & 78 &  20        & 800       \\ \hline
			Slip Detection & 0.15              & 0.001                   & 150 & 78 &  2         & 100       \\ \hline
	\end{tabular}}
	\label{t1}
\end{table}
\subsection{Datasets}
In this paper, we use the datasets collected by NeuTouch~\cite{taunyazov2020event}. Specifically, three datasets are collected, including ``Objects'' and ``Containers'' for event-driven tactile object recognition and ``Slip Detection'' for event-driven slip detection. Unlike ``Objects'' only requiring models to determine the type of objects being handled, ``Containers'' asks models about the type of containers being handled and the amount of liquid (0\%, 25\%, 50\%, 75\%, 100\%) held within. Thus, ``Containers'' is more challenging for event-driven tactile object recognition. Moreover, the task of event-driven slip detection is also challenging since it requires models to detect the rotational slip within a short time, like 0.15s for ``Slip Detection''. We summarize the dataset statistics in Table~\ref{t1}, where TD is for time durations, SSR is for spike sampling rates, $T$=TD / SSR is the total time length, $N$ is the total number of taxels (each tactile sensor has 39 taxels and two tactile sensors are used), and $K$ is the number of classes. We split the data into a training set (80\%) and a test set (20\%) with an equal class distribution in the experiments. We repeat each experiment for five rounds and report the average accuracy. 

\begin{table}[]
	\begin{threeparttable}
	\caption{Accuracies on the event-driven tactile learning datasets}
	\setlength{\tabcolsep}{2.3mm}{
		\begin{tabular}{lcccc}
			\hline
			Method       & Type & Objects       & Containers    & Slip Detection \\ \hline
			Tactile-SNN~\cite{taunyazov2020event}       & SNN  & 0.75          & \hspace{1.5mm}0.57*          & \hspace{1.5mm}0.82*           \\ \hline
			TactileSGNet~\cite{gu2020tactilesgnet} & SNN  & 0.79          & 0.58          & 0.97           \\ \hline
			GRU-MLP~\cite{taunyazov2020event}      & ANN  & 0.72          & \hspace{1.5mm}0.46*          & \hspace{1.5mm}0.87*           \\ \hline
			CNN-3D~\cite{taunyazov2020event}       & ANN  & 0.90          & \hspace{1.5mm}0.67*          & \hspace{1.5mm}0.44*           \\ \hline
			Ours-Hybrid  & SNN  & \textbf{0.91} & \textbf{0.86} & \textbf{1.0}   \\ \hline
	\end{tabular}}\label{t2}
\begin{tablenotes}
	\small
	\item *These values come from the paper~\cite{taunyazov2020event}. And the best performance is in bold.
\end{tablenotes}
\end{threeparttable}
\end{table}

\subsection{Models}
We compare our models with the state-of-the-art SNN methods for event-driven tactile learning, including Tactile-SNN~\cite{taunyazov2020event} and TactileSGNet~\cite{gu2020tactilesgnet}. Tactile-SNN employs \textbf{TSRM neurons} as the building blocks, and the network structure of Tactile-SNN is Input-SFc0-SFc1. While TactileSGNet utilizes \textbf{LIF neurons} as the building blocks and proposes the spiking graph neural network (SGNet). The network structure of TactileSGNet is Input-SGNet-SFc1-SFc2-SFc3. We also compare our models against conventional deep learning, specifically Gated Recurrent Units (GRUs)~\cite{cho2014learning} with Multi-layer Perceptrons (MLPs) and 3D convolutional neural networks~\cite{gandarias2019active}. The network structure of GRU-MLP is Input-GRU-MLP, where MLP is only utilized at the final time step. And the network structure of CNN-3D is Input-3D\_CNN1-3D\_CNN2-Fc1.

\begin{table}[]
	\caption{Accuracies for ablation studies}
	\setlength{\tabcolsep}{1.3mm}{
		\begin{tabular}{lcccc}
			\hline
			Method               & Type & Objects & Containers & Slip Detection \\ \hline
			Tactile-SNN~\cite{taunyazov2020event}               & SNN  & 0.75    & 0.57       & 0.82           \\ \hline
			Ours-Location Tactile-SNN & SNN  & 0.89    & 0.88       & 0.82           \\ \hline
			Ours-Hybrid $\lambda=1$          & SNN  & 0.91    & 0.86       & 1.0            \\ \hline
			Ours-Hybrid $\lambda=0.5$         & SNN  & 0.92    & 0.89       & 0.98           \\ \hline
			Ours-Hybrid-loop  & SNN  & 0.91    & 0.86       & 1.0            \\ \hline
			Ours-Hybrid-arch  & SNN  & 0.91    & 0.86       & 0.99           \\ \hline
			Ours-Hybrid-whorl & SNN  & 0.92    & 0.86       & 0.98           \\ \hline
	\end{tabular}}\label{t3}
\end{table}
\subsection{Performance and Analysis}
\subsubsection{Basic Performance}
Table~\ref{t2} presents the test accuracies on the three datasets. We observe that our hybrid model significantly outperforms the state-of-the-art SNNs. Moreover, figure~\ref{profile} shows the training and testing profiles for Tactile-SNN and our hybrid model. From this figure, we can see that our model converges faster and attains the lower loss and the higher accuracy compared to Tactile-SNN. The reason why our model is superior to other SNNs could be two-fold: (1) different from state-of-the-art SNNs that only extract features with existing spiking neurons, our model employs an SNN with location spiking neurons to extract features in a novel way; (2) our model fuses SNN\_TSRM and SNN\_LSRM to better capture complex spatio-temporal dependencies in the data. We also compare our model with ANNs, which provide fair comparison baselines for fully ANN architectures since they employ similar lightsome network architectures as ours. From Table~\ref{t2}, we find out that our model outperforms ANNs on the three tasks, which might be because our model is more compatible with various kinds of event-based data and better maintains the sparsity to prevent overfitting.

\subsubsection{Ablation Studies}
To examine the effectiveness of each component in the hybrid model, we \textbf{seperately train} SNN\_TSRM (which is exactly Tactile-SNN) and SNN\_LSRM (which is referred to as Location Tactile-SNN). From Table~\ref{t3}, we surprisingly find out that Location Tactile-SNN significantly surpasses Tactile-SNN on the datasets for event-driven tactile object recognition and provides comparable performance on the event-driven slip detection. The reason for this could be two-fold: (1) the time durations of event-driven tactile object recognition datasets are longer than that of ``Slip Detection'', and Location Tactile-SNN is good at capturing the mid-and-long term dependencies in these object recognition datasets; (2) like Tactile-SNN, Location Tactile-SNN can still capture the spatial dependencies in the event-driven tactile data (``Slip Detection'') due to the spatial recurrent neuronal dynamics of location spiking neurons. Furthermore, we examine the sensitivities of $\lambda$ in Eq.(\ref{e8}) and location orders. From Table~\ref{t3}, we notice the results of related models are close, proving that the $\lambda$ tuning and location orders do not significantly impact the task performance.


\subsubsection{Confusion Matrices}
\begin{figure}
	\includegraphics[width=\linewidth]{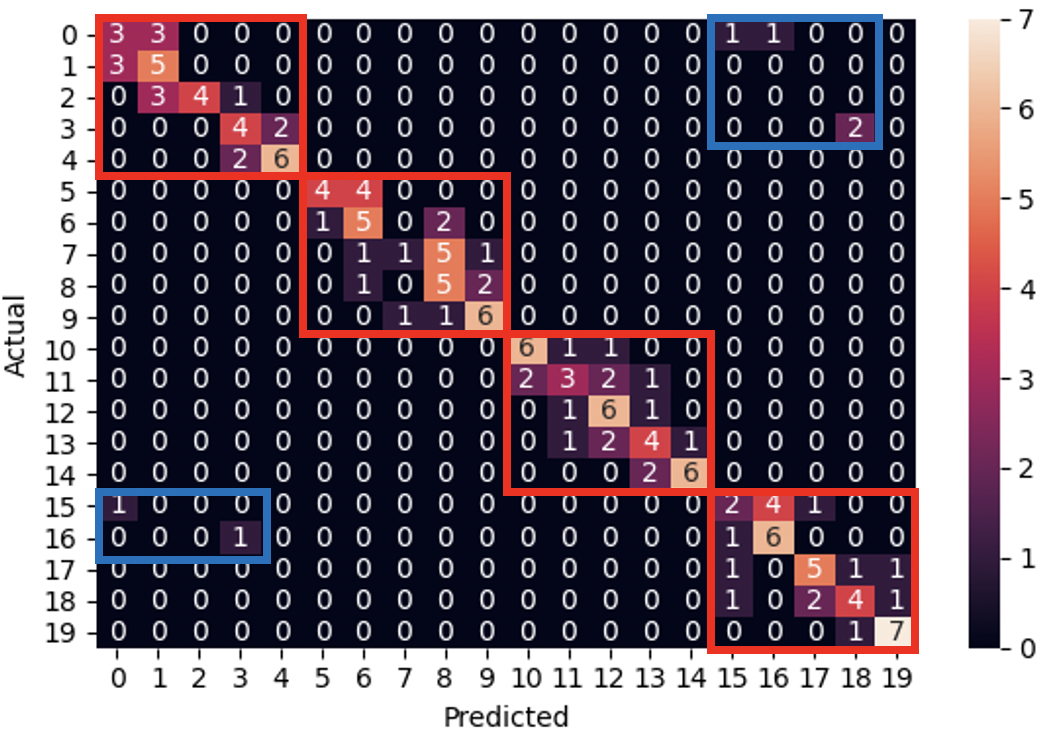}
	\caption{The confusion matrix of Tactile-SNN on ``Containers''.}
	\label{cm1}
\end{figure}
\begin{figure}
	\includegraphics[width=\linewidth]{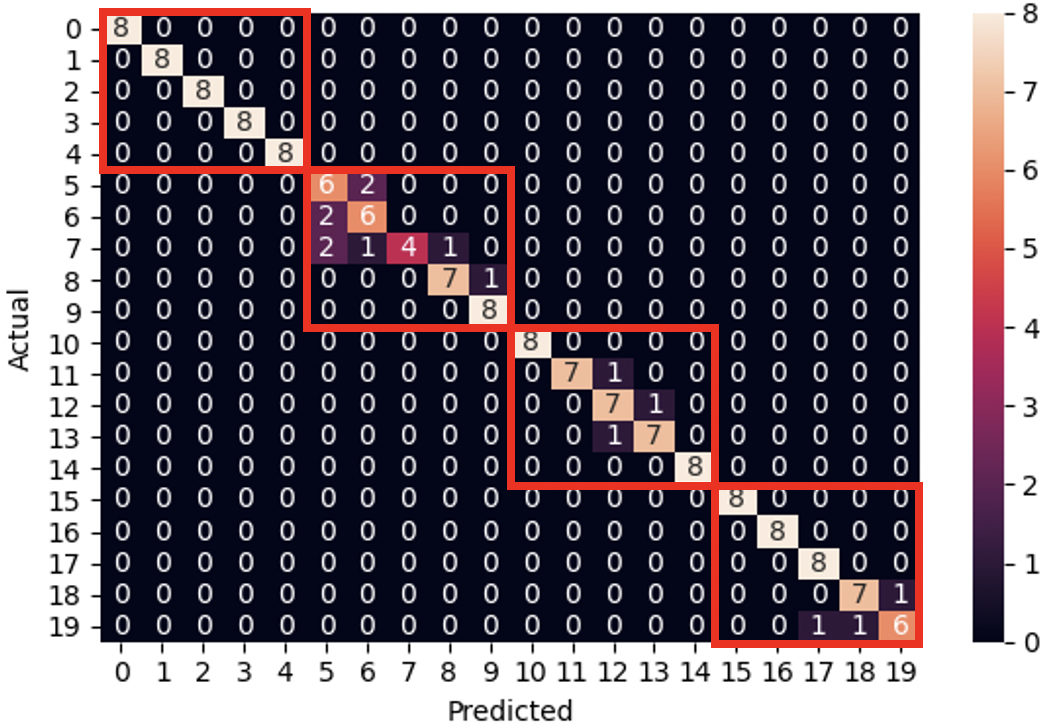}
	\caption{The confusion matrix of Ours-Hybrid on ``Containers''.}
	\label{cm2}
\end{figure}
We calculate the confusion matrices of Tactile-SNN (Fig.~\ref{cm1}) and our hybrid model (Fig.~\ref{cm2}) on ``Containers'' since it is a more challenging event-driven tactile object recognition dataset. From the two figures, we can see that our hybrid model can perfectly distinguish the different containers. Each red box in the figures represents a type of container, and each blue box in the figures represents the container misclassification. Moreover, compared to Tactile-SNN, we observe that our model can recognize the container fullness with a higher accuracy since the misclassification number in each red box is fewer for our model.

\subsubsection{Timestep-wise Inference}
We evaluate the timestep-wise inference performance of the hybrid model and validate the contributions of the two components in it. Moreover, we propose a time-weighted hybrid model to better balance the two components' contributions and achieve the better overall performance. Figure~\ref{exp}(a), \ref{exp}(b), and \ref{exp}(c) show the timestep-wise inference accuracies (\%) of SNN\_TSRM, SNN\_LSRM, the hybrid model, and the time-weighted hybrid model on the three datasets. Specifically, the output of the time-weighted hybrid model at time $t$ is 
\begin{equation}
	\label{e9}
	\begin{split}
	O_{tw}(t) = concate&nate((1-\omega) * O1(t), \omega * O2(t)), \\
    \omega &= \frac{1}{1+e^{-\psi*(\frac{t}{T}-1) }},
	\end{split}
\end{equation}
where the hyperparameter $\psi$ controls the balance between SNN\_TSRM's contribution and SNN\_LSRM's contribution and $T$ is the total time length. From the figures, we can see that SNN\_TSRM has good ``early'' accuracies on the three tasks since it well captures the spatial dependencies with the help of Eq.~(\ref{e5}). However, its accuracies do not improve too much at the later stage since it does not sufficiently capture the temporal dependencies. In contrast, SNN\_LSRM has fair ``early'' accuracies, while its accuracies jump a lot at the later stage since it models the temporal dependencies in Eq.~(\ref{e6}). The hybrid model adopts the advantages of these two components and extracts spatio-temporal features from various views, which enables it to have a better overall performance. Furthermore, after employing the time-weighted output and shifting more weights to SNN\_TSRM at the early stage, the time-weighted hybrid model can have a good ``early'' accuracy as well as an excellent ``final'' accuracy.
\begin{figure*}
	\includegraphics[width=\linewidth,height=0.58\linewidth]{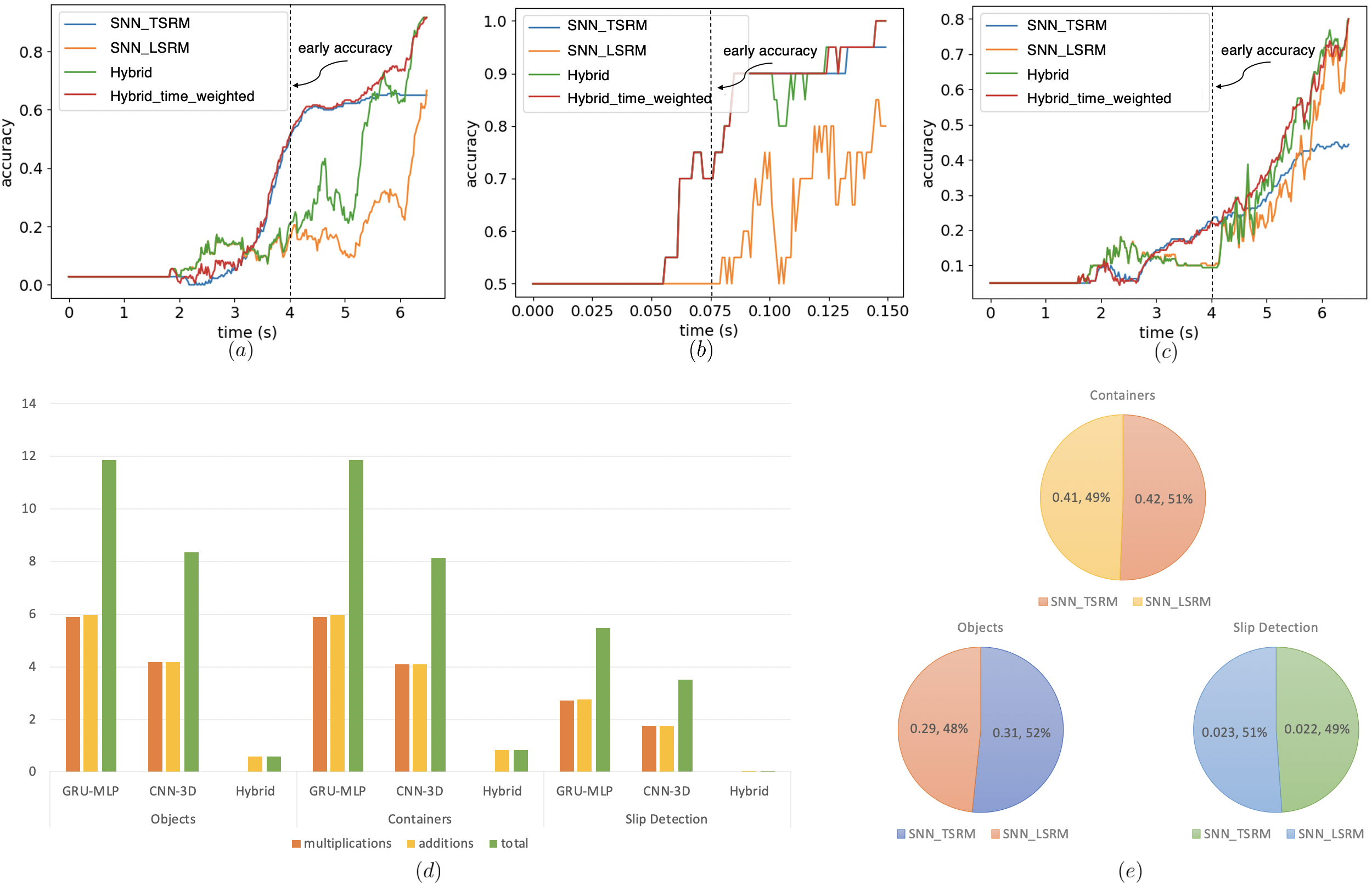}
	\caption{The timestep-wise inference (Alg.~\ref{alg:timestep}) for SNN\_TSRM -- $O1$, SNN\_LSRM -- $O2$, the hybrid model -- $O$, and the time-weighted hybrid model -- $O_{tw}$ on (a) ``Objects'', (b) ``Slip Detection'', (c) ``Containers''. Note that we use the same event sequences as~\cite{taunyazov2020event} and the first spike occurs at around 2.0s for ``Objects'' and ``Containers''. (d) Operation cost ($\times 10^6$) comparison between ANNs and Ours-Hybrid on the benchmark datasets. (e) Operation cost ($\times 10^6$) and percentage comparison between SNN\_TSRM and SNN\_LSRM on the benchmark datasets.}
	\label{exp}
\end{figure*}
\subsection{Energy Efficiency}
\begin{table}[]
	\caption{Operation cost ($\times 10^6$) and compression ratio on benchmark datasets}
	\setlength{\tabcolsep}{2.5mm}{
	\begin{tabular}{lccc}
		\hline
		Method      & Objects                               & Containers                             & Slip Detection                          \\ \hline
		GRU-MLP     & 11.87                                 & 11.87                                  & 5.48                                    \\ \hline
		CNN-3D      & 8.34                                  & 8.14                                   & 3.5                                     \\ \hline
		Ours-Hybrid & 0.60                                  & 0.83                                   & 0.045                                   \\ \hline
		Ratio       & 9.80$\sim$14.30$\times$ & 13.90$\sim$19.78$\times$ & 77.78$\sim$121.78$\times$ \\ \hline
	\end{tabular}}\label{t4}
\end{table}
Following the estimation method in \cite{ijcai2021-441, lee2020spike}\footnote{We consider the computational costs in feature matrix transformation.}, we estimate the computational costs of the hybrid model and ANNs on the three datasets. As shown in Fig.~\ref{exp}(d), the hybrid model has no multiplication operations and achieves far fewer addition operations than ANN models on the three datasets. Moreover, based on Table~\ref{t4}, the compression ratio of total operations (ANNs Opts.~/~Ours Opts.) is between $\mathbf{9.80\times}$
and $\mathbf{121.78\times}$. These results are consistent with the fact that the sparse spike communication and event-driven computation underlie the efficiency advantage of SNNs and demonstrate the potentials of our model on neuromorphic hardware. 
We further compare the costs of SNN\_TSRM and SNN\_LSRM on the benchmark datasets. From Fig.~\ref{exp}(e), we can see that the cost of SNN\_LSRM is almost equal to that of SNN\_TSRM on each dataset, which shows that the location spiking neurons have the similar energy efficiency compared to existing spiking neurons. Such high-efficiency benefits make location spiking neurons a perfect fit for neuromorphic hardware.
\section{Discussion and Conclusion}\label{sec:conclude}
This paper proposes a novel neuron model -- ``location spiking neuron'' and introduces the spatial recurrent neuronal dynamics of LSRM neurons. We believe the idea of such location spiking neurons can be applied to other existing spiking neuron models like LIF neurons and strengthen their feature representation abilities. Moreover, we think the location spiking neurons can build more complicated models to further boost the event-driven tactile learning performance. For example, we can develop a spiking graph neural network with location spiking neurons and combine it with ~\cite{gu2020tactilesgnet} to better serve event-driven tactile learning tasks. Furthermore, besides event-driven tactile learning, we can apply the models with location spiking neurons to other event-driven learning fields, like event-based vision or event-driven audio sensing. By analyzing the applications in these fields, we can further understand the strengths and weaknesses of this new neuron.


In this work, we propose location spiking neurons and demonstrate the dynamics of LSRM neurons. By exploiting the LSRM neurons, we develop several models for event-driven tactile learning to sufficiently capture the complex spatio-temporal dependencies. The experimental results on three datasets demonstrate the extraordinary performance and high energy efficiency of our models and location spiking neurons. This further unlocks their potential on neuromorphic hardware. Overall, this work sheds new light on SNN representation learning and event-driven learning, which may facilitate the understanding of advanced cognitive intelligence.

\bibliographystyle{IEEEbib}
\bibliography{strings,refs}

\end{document}